\documentclass{article} 
\usepackage{iclr2027_conference,times}

\usepackage{amsmath,amsfonts,bm}

\def\eqref#1{equation~\ref{#1}}

\def\1{\bm{1}}

\DeclareMathAlphabet{\mathsfit}{\encodingdefault}{\sfdefault}{m}{sl}
\SetMathAlphabet{\mathsfit}{bold}{\encodingdefault}{\sfdefault}{bx}{n}

\usepackage{hyperref}
\usepackage{url}
\usepackage{booktabs}
\usepackage{multirow}
\usepackage[table]{xcolor}
\usepackage{graphicx}
\usepackage{enumitem}
\usepackage{amsmath}
\usepackage{amssymb}
\usepackage{wrapfig}
\usepackage{algorithm}
\usepackage{algpseudocode}
\usepackage{caption}
\usepackage{xspace}
\newcommand{\method}{\textsc{B-OPSD}\xspace}
\usepackage{tcolorbox}
\tcbuselibrary{breakable,skins}

\usepackage{xcolor}
\algrenewcommand\algorithmiccomment[1]{%
  \hfill\textcolor{black!55}{$\triangleright$~#1}%
}

\definecolor{promptborder}{gray}{0.68}
\definecolor{promptheader}{gray}{0.68}
\definecolor{promptbody}{gray}{0.97}

\newtcolorbox{promptbox}[1]{
  enhanced,
  colback=promptbody,
  colframe=promptborder,
  colbacktitle=promptheader,
  coltitle=white,
  title={#1},
  fonttitle=\bfseries,
  fontupper=\footnotesize\ttfamily\raggedright,
  boxrule=0.6pt,
  arc=2pt,
  outer arc=2pt,
  left=7pt,
  right=7pt,
  top=6pt,
  bottom=6pt,
  toptitle=4pt,
  bottomtitle=4pt,
  before skip=6pt,
  after skip=6pt
}

\title{Train Ahead, Distill Back: Bootstrapping On-Policy Self-Distillation for Large Language Models}

\author{
Zheng Zhang\textsuperscript{1,2}\thanks{Equal contribution.}\hspace{0.4em},
Xinyue Tan\textsuperscript{1}\footnotemark[1]\hspace{0.4em},
Lufei Li\textsuperscript{1},
Xinyi Zhang\textsuperscript{3},
Yexin Li\textsuperscript{2}\footnotemark[2]\hspace{0.4em},
Kan Ren\textsuperscript{1}\thanks{Corresponding authors.} \\
\textsuperscript{1}School of Information Science and Technology, ShanghaiTech University \\
\textsuperscript{2}State Key Laboratory of General Artificial Intelligence, BIGAI \\
\textsuperscript{3}Hefei University of Technology \\
\texttt{\{zhangzheng2024,tanxy2024,renkan\}@shanghaitech.edu.cn}
}

\iclrfinalcopy 
\begin{document}

\maketitle
\fancyhead[L]{}

\begin{abstract}
On-policy self-distillation (OPSD) improves large language models by letting a self-teacher with privileged information provide dense token-level supervision on the model's own trajectories.
Yet existing methods typically construct the self-teacher from the current, initial, or slowly averaged policy state, leaving the quality of supervision constrained by the teacher's ability to exploit privileged information.
We ask whether the \textit{model's own optimization} progress can instead be \textit{recycled} into a stronger self-teacher.    
In this paper, we introduce Bootstrapped On-Policy Self-Distillation (\method), which temporarily trains the policy ahead to obtain a future teacher, restores the student to the original policy state, and then uses the future teacher to supervise the restarted student.
The future teacher improves supervision in two complementary ways, it can generate more reliable privileged trajectories and, conditioned on them, provide more informative token-level targets along the restarted student's on-policy trajectories.
Experiments on mathematical reasoning with Qwen3-4B and Qwen3-8B show consistent improvements over standard OPSD in both settings, including gains from 27.50 to 41.30 and from 48.80 to 64.44 in the rollout-privileged setting.
Our findings point to a broader principle for self-improving models that future learning progress can be distilled backward, preserving acquired knowledge while bootstrapping beyond the optimization state that produced it.
\end{abstract}

\section{Introduction}
On-policy self-distillation (OPSD) has emerged as a promising paradigm for improving the reasoning capabilities of large language models (LLMs) without relying on a stronger external teacher \citep{penaloza2026privileged,shenfeld2026selfdistillation,zhang2026learning}.
It uses the same model as both teacher and student, with the teacher additionally conditioned on privileged information such as reference solutions, final answers or successful rollouts \citep{gkountouras2026consensus, zhao2026selfdistilled}.
This privileged context enables the self-teacher to provide dense token-level supervision along the student's own trajectories.

However, the quality of OPSD depends not only on what privileged information is available, but also on how effectively the self-teacher can exploit it.
Existing methods have explored increasingly informative privileged signals, including final answers and model-generated rollouts \citep{hubotter2026reinforcement,li2026unifying,tian2026pbsd}.
Yet their self-teachers are typically constructed from the initial, current, or slowly averaged policy state \citep{shenfeld2026selfdistillation,hubotter2026reinforcement}.
As a result, even informative privileged context is interpreted by a teacher whose capability remains tied to roughly the same stage of optimization as the learner.
This motivates a complementary question.
\textit{Can a model recycle its own optimization progress into stronger supervision for an earlier version of itself?}

We introduce Bootstrapped On-Policy Self-Distillation (\method), a train-ahead-and-distill-back framework that turns temporary optimization progress into supervision.
Starting from the initial policy, we perform a short sequence of temporary OPSD updates to obtain a future policy, which we use as the future teacher.
We then restore the student to its original state and distill from this improved self.
The future teacher improves both the source and use of privileged information.
It can generate more reliable privileged trajectories and, conditioned on them, provide more informative token-level targets along the restarted student's own trajectories.
The student therefore remains on-policy with respect to its restarted state while benefiting from knowledge acquired later in optimization.
Rather than simply continuing from a stronger checkpoint, \method separates the progress captured by that checkpoint from the optimization state that produced it.

We instantiate \method in both answer-available and answer-free settings.
When gold answers are available, verified future-policy trajectories selectively activate future-teacher supervision.
Without gold answers, \method uses agreement among model generations and queries the future teacher when the restarted student's own consensus is insufficient.
This novel selective mechanism applies stronger supervision where the current policy provides a less reliable learning signal, while retaining standard OPSD elsewhere.

Experiments on three widely adopted public benchmarks with two open-source LLMs show consistent and substantial improvements over standard OPSD.
Controlled analyses show that both the future teacher and its compatible privileged trajectories contribute to the improvement, and that future supervision is most effective early in restarted training.
Comparisons of matched training budget further show that our proposed \method is more effective in learning and the gains are not solely by the additional updates used to construct the future teacher.
As shown in optimization-landscape analysis in Section \ref{sec:restart-landscape}, distilling back from the future can redirect the restarted student toward higher-reward regions than simply continuing optimization from the future checkpoint.
Together, these results suggest that optimization progress can serve not only as updated parameters, but also as a reusable source of supervision for self-improvement.

Our main contributions are as follows:
\begin{itemize}[leftmargin=1.0em, labelsep=0.3em, nosep]
    \item We propose \method, a train-ahead-and-distill-back framework that converts optimization progress into supervision for an earlier policy state while preserving on-policy learning for the restarted student.
    \item We show that future supervision improves both the source and interpretation of privileged information, enabling selective self-distillation with and without reference answers.
    \item Across two model scales and three mathematical reasoning benchmarks, \method consistently improves standard OPSD, while controlled analyses show that the gains arise from how future progress is reused as supervision rather than from additional optimization alone.
\end{itemize}

\begin{figure}[t]
    \centering
    \includegraphics[width=\linewidth]{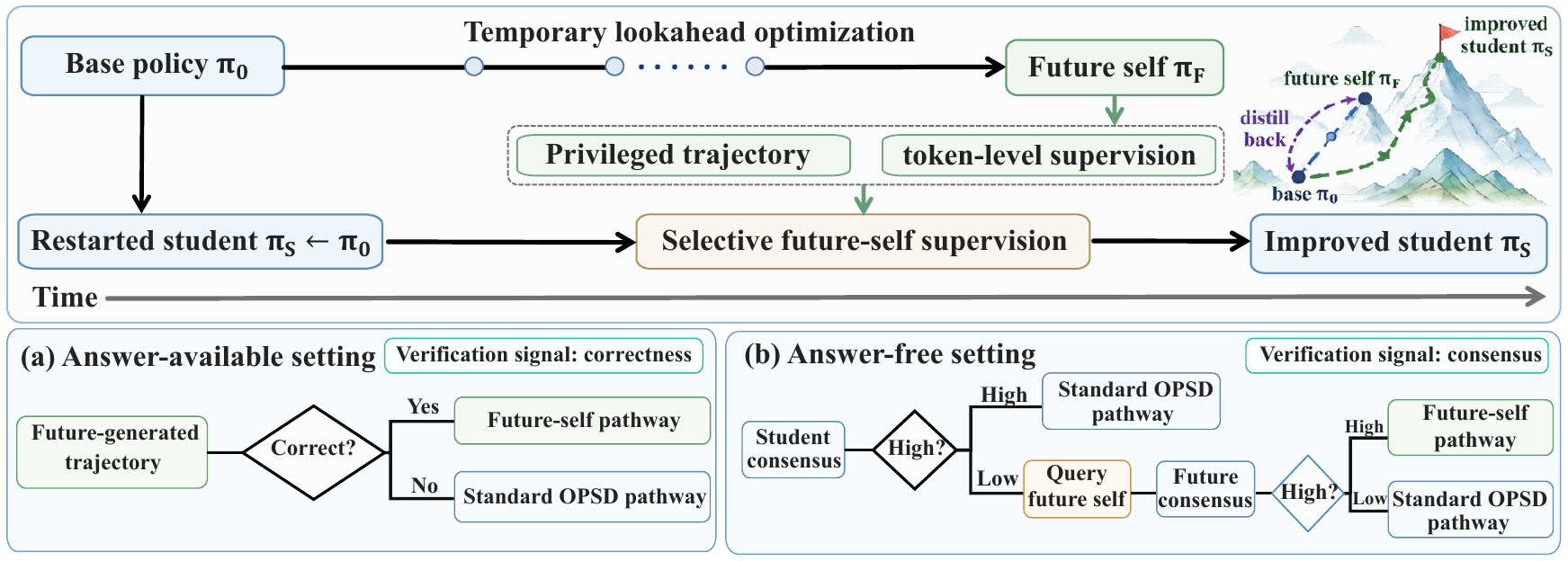}
    \caption{Overview of \method. We first train the base policy $\pi_0$ forward to obtain a future self $\pi_F$, then restart the student from $\pi_0$ and distill from $\pi_F$ when future supervision is reliable.
The future self supplies both privileged trajectories and token-level targets. In the answer-available setting, reliability is determined by correctness; in the answer-free setting, it is determined by student and future consensus.
    }
    
    \label{fig:framework}
    \vspace{-1em}
\end{figure}

\section{Related Work}
\paragraph{On-Policy Self-Distillation.}
OPSD enables a model to serve as both teacher and student under different contexts \citep{zhang2026latent,li2026policy,yang2026ogls,narozniak2026distilling}. 
The student generates trajectories, while the teacher additionally conditions on privileged information and provides token-level supervision along these trajectories.
The two key components of OPSD are the privileged information and the model serving as the teacher.
Several types of privileged information are commonly used:
final answer \citep{tian2026pbsd,yang2026self},
reference solution \citep{zhao2026selfdistilled,shen2026purified},
model-generated rollouts \citep{hubotter2026reinforcement,li2026unifying},
structured guidance \citep{xia2026enhancing,gu2026rethinking}.
Regarding the choice of teacher model, common designs include a frozen copy of the initial student \citep{fu2026reducing,yang2026adaptive}, an exponential moving average (EMA) of the student parameters \citep{zhang2026sdpo,jin2026unisd,hubotter2026reinforcement}, and a teacher that shares the current student’s parameters exactly \citep{stein2026gates,yang2026does}.

\paragraph{Learning across Optimization States.}
Model optimization produces a sequence of intermediate states that may contain complementary supervisory information.
Most existing methods \citep{yang2019snapshot,laine2017temporal,zhang2024scac,wang2022efficient,wang2022bootstrapped} reuse states from earlier stages of optimization to guide subsequent learning.
\citet{laine2017temporal} and \citet{tarvainen2017mean} aggregate predictions or parameters from historical model states to construct more stable training targets.
\citet{wei2026gtr} merges multiple checkpoints into a consolidated teacher that guides subsequent policy training.
Several lines of work \citep{flennerhag2022bootstrapped,du2022learning,oh2025future} have explored using later optimization states to guide earlier stages of learning.
For example, 
\citet{qin2026near} obtains a stronger policy by continuing RLVR and injects its verifier-confirmed trajectories into the earlier policy's rollout groups for RL training. 
\citet{li2026rise} extrapolates an RLVR-induced policy update to construct a synthetic teacher, which is then distilled back into the policy.
Nevertheless, how to use future policy states as teachers remains underexplored, particularly in OPSD. Our method uses the future teacher to provide stronger trajectories as privileged information and improved token-level targets.

\section{Preliminary}
\paragraph{On-Policy Self-Distillation.}

Given a problem $x \sim \mathcal{D}$, the student samples a trajectory
$y \sim \pi_\theta(\cdot \mid x)$.
The same model serves as a self-teacher by additionally conditioning on
privileged information $z$.
OPSD aligns the student's next-token distribution with the teacher's
along student-generated trajectories using the forward KL divergence:
\begin{equation}
\label{eq:opsd}
\mathcal{L}_{\mathrm{OPSD}}(\theta)
=
\mathbb{E}_{x \sim \mathcal{D},\, y \sim \pi_\theta(\cdot \mid x)}
\left[
\frac{1}{|y|}
\sum_{t=1}^{|y|}
D_{\mathrm{KL}}
\left(
\operatorname{sg}\!\left[
\pi_\theta(\cdot \mid x, z, y_{<t})
\right]
\,\middle\|\,
\pi_\theta(\cdot \mid x, y_{<t})
\right)
\right].
\end{equation}
Here, $y_{<t}$ denotes the trajectory prefix, and
$\operatorname{sg}[\cdot]$ is the stop-gradient operator, which prevents
gradients from propagating through the teacher branch.

\section{Methodology}

\subsection{Learning from a Future Self}
\label{sec:future_self}
We introduce \method, which converts temporary OPSD progress into supervision for an earlier policy state. Let $\pi_0$ denote the initial policy and $\mathcal{A}$ an OPSD update. We perform $K$ lookahead updates starting from $\pi_0$ to obtain a \emph{future self},
\begin{equation}
    \pi_F = \mathcal{A}^{K}(\pi_0).
\end{equation}

We then freeze $\pi_F$ and restart the student from $\pi_S=\pi_0$.
During restarted training, the student generates the on-policy trajectories along which distillation is performed. 
For each update, we select one of two supervision pathways, each pairing a teacher with a trajectory used as privileged information. 
(i) In the \textit{future-self pathway}, $\pi_F$ generates the privileged trajectory and, conditioned on it, provides token-level targets. 
(ii) In the \textit{standard OPSD pathway}, the frozen $\pi_0$ provides these targets, conditioned on privileged information available in the training setting.

Once the teacher and privileged trajectory are selected, we use the following token-level distillation objective to update the student in either pathway. Let $y_S$ denote the student trajectory, $y_P$ the privileged trajectory, and $\pi_T$ the frozen teacher:
\begin{equation}
\label{eq:bootstrapped_opsd}
\mathcal{L}_{\mathrm{distill}}(\pi_S,\pi_T;y_S,y_P)
=
\frac{1}{|y_S|}
\sum_{t=1}^{|y_S|}
D_{\mathrm{KL}}\left(
\operatorname{sg}\!\left[
\pi_T(\cdot\mid x,y_P,y_{S,<t})
\right]
\,\middle\|\,
\pi_S(\cdot\mid x,y_{S,<t})
\right),
\end{equation}
where $\pi_T$ is the frozen teacher selected by the pathway.

We study two settings that differ in whether gold answers are available during training: an answer-available setting with gold answers for verifying trajectories, and an answer-free setting that relies on agreement among generated answers. 
In the answer-available setting, we use the future-self pathway when a future-generated trajectory is verified as correct. 
In the answer-free setting, we use the future-self pathway when the restarted student's answers show insufficient agreement but the future policy's answers reach the agreement threshold. 
Otherwise, we follow standard OPSD. 
In both settings, distillation remains on-policy with respect to the restarted student.

\subsection{Answer-available setting.}
\label{sec:answer_available}
We first consider a setting in which gold answers are available for trajectory verification and gold reasoning traces provide a fallback source of privileged information. 
The overall procedure consists of the following two stages, with the complete workflow detailed in Algorithm~\ref{alg:lookahead_opsd_answer}.

\textbf{Stage I: Construct the future self.}
To construct the future self, we initialize the policy from $\pi_0$ and perform $K$ OPSD updates. 
At each update, an incorrect on-policy trajectory generated by the student serves as the student trajectory, while a frozen copy of the initial policy serves as the teacher and provides target token distributions conditioned on additional privileged information.
The privileged information is selected according to a self-generated-first principle: a correct student-generated trajectory is used whenever available, with the gold reasoning trace serving as a fallback.
This lookahead stage yields the future policy $\pi_F$, which represents an improved policy state reached after $K$ self-distillation updates from $\pi_0$.

\textbf{Stage II: Restart student and learn from the future self.}
After constructing $\pi_F$, we freeze it and reinitialize the student as $\pi_S=\pi_0$. For each problem, the student generates a set of on-policy trajectories, from which an incorrect one is selected as the student trajectory. The future policy then generates a candidate reasoning trajectory whose correctness determines the source of supervision.
When this trajectory is correct, we take the future-self pathway: the frozen $\pi_F$, conditioned on the future-generated trajectory, provides the target token distributions.
Otherwise, we take the standard OPSD pathway, using the frozen $\pi_0$ as the teacher and conditioning it on either a correct student-generated trajectory or, when none is available, the gold reasoning trace.

\subsection{Answer-Free Setting}
We next consider an answer-free setting in which training has access only to problem statements, without gold answers or reasoning traces. 
In this setting, we use agreement among model-generated answers as a proxy for correctness.
The overall procedure consists of two stages, with the complete workflow detailed in Algorithm~\ref{alg:lookahead_opsd_answer_free}.

\textbf{Stage I: Construct the future self.}
We initialize the student from $\pi_0$ and perform $K$ OPSD updates.
For each problem, the student samples $N$ on-policy trajectories and derives a pseudo-label $\hat{a}$ by majority vote over their final answers. 
A trajectory that disagrees with $\hat{a}$ is selected as the student trajectory for
distillation, while a trajectory supporting $\hat{a}$ serves as privileged
information. 
A frozen copy of the initial policy $\pi_0$, conditioned on this privileged trajectory, serves as the teacher and provides the target token distributions. 
Thus, in the absence of gold answers, the lookahead stage constructs its training signal entirely from agreement among the student's own rollouts. 
The policy obtained after these $K$ updates is denoted by $\pi_F$ and serves as the future self.

\textbf{Stage II: Restart student and learn from the future self.}
After constructing $\pi_F$, we freeze it and reinitialize the student as $\pi_S=\pi_0$.
For each problem, the restarted student samples $N$ on-policy trajectories and takes their most frequent final answer as a pseudo-label $\hat{a}_S$.
If more than half of these trajectories support $\hat{a}_S$, we follow the standard OPSD pathway: the frozen $\pi_0$ serves as the teacher, conditioned on a student-generated trajectory supporting $\hat{a}_S$.
Otherwise, we sample $N$ trajectories from $\pi_F$ and take their most frequent final answer as a pseudo-label $\hat{a}_F$.
If more than half of the future trajectories support $\hat{a}_F$, we take the future-self pathway: $\pi_F$ serves as the teacher, conditioned on a future-generated trajectory supporting $\hat{a}_F$.
If this agreement threshold is not met, we fall back to the standard OPSD pathway using $\hat{a}_S$ and a student-generated trajectory supporting it.
In all cases, distillation is performed along a restarted student's trajectory whose answer disagrees with the pseudo-label used for supervision.

\section{Experiments}
\subsection{Experimental Settings}
\paragraph{Models and Baselines.}
We use Qwen3-4B and Qwen3-8B~\citep{yang2025qwen3} as the policy models, with thinking disabled throughout training and evaluation. 
We use vanilla OPSD~\citep{zhao2026selfdistilled} as the primary baseline.
Unless otherwise specified, all subsequent analysis experiments use Qwen3-4B as the policy model.

\paragraph{Implementation Details.}
We train on the OPSD split of OpenThoughts-Math-30K~\citep{guha2026openthoughts}, which contains 29,434 problems. Each update samples 32 problems. We consider both single-rollout ($n=1$) and multi-rollout ($n=8$) settings, corresponding to 32 and 256 trajectories per update, respectively. Responses are sampled with a temperature of 1.1, top-$p$ of 0.95, and top-$k$ of 20, with a maximum length of 4,096 tokens. We optimize LoRA adapters with rank 64, scaling factor 128, and a learning rate of $5\times10^{-6}$.

\paragraph{Evaluation.}
We evaluate on AIME 2024 \citep{maa2024aime}, AIME 2025 \citep{maa2025aime}, and HMMT February 2025 \citep{hmmt2025}. 
For each problem, we sample 12 responses with a temperature of 1.0, top-$p$ of 0.8, and no top-$k$ truncation. We set the maximum response length to 38,912 tokens. We report three metrics: \textsc{Avg@12}, the average accuracy over 12 sampled responses; \textsc{Pass@12}, the fraction of problems for which at least one of the 12 responses is correct; and \textsc{Maj@12}, the accuracy of the majority-voted answer. All metrics are reported as percentages.

\subsection{Main Results}

\begin{table}[tbp]
\centering
\small
\setlength{\tabcolsep}{4.5pt}
\renewcommand{\arraystretch}{1.10}
\caption{Main results comparison across mathematical reasoning benchmarks under
the answer-available setting. OPSD follows the original reference-solution
setting, while OPSD$^\dagger$ denotes our rollout-based variant with
student-rollout PI. The best results are highlighted in \textbf{bold}.
}
\label{tab:answer-supervised}
\begin{tabular}{lllcrrrr}
\toprule
Model & Method & Privileged Information & $n$
& AIME24 & AIME25 & HMMT25 & Avg. \\
\midrule

\multirow{5}{*}{\textit{Qwen3-8B}}
& Base & -- & -- & 30.55 & 20.83 & 11.11 & 20.83 \\

\cmidrule(lr){2-8}

& OPSD & reference solution & 1
& 52.50 & 38.89 & 20.28 & 37.22 \\

& \cellcolor{black!6}\method
& \cellcolor{black!6}reference solution
& \cellcolor{black!6}1
& \cellcolor{black!6}\textbf{58.61}
& \cellcolor{black!6}\textbf{46.39}
& \cellcolor{black!6}\textbf{27.50}
& \cellcolor{black!6}\textbf{44.17} \\

\cmidrule(lr){2-8}

& OPSD$^\dagger$ & Student rollout & 8
& 61.67 & 51.94 & 32.78 & 48.80 \\

& \cellcolor{black!12}\method
& \cellcolor{black!12}Future rollout
& \cellcolor{black!12}8
& \cellcolor{black!12}\textbf{76.94}
& \cellcolor{black!12}\textbf{66.94}
& \cellcolor{black!12}\textbf{49.44}
& \cellcolor{black!12}\textbf{64.44} \\

\midrule

\multirow{5}{*}{\textit{Qwen3-4B}}
& Base & -- & -- & 20.00 & 19.17 & 10.83 & 16.67 \\

\cmidrule(lr){2-8}

& OPSD & reference solution & 1
& 32.78 & 24.72 & 16.39 & 24.62 \\

& \cellcolor{black!6}\method
& \cellcolor{black!6}reference solution
& \cellcolor{black!6}1
& \cellcolor{black!6}\textbf{36.67}
& \cellcolor{black!6}\textbf{27.22}
& \cellcolor{black!6}\textbf{18.33}
& \cellcolor{black!6}\textbf{27.41} \\

\cmidrule(lr){2-8}

& OPSD$^\dagger$ & Student rollout & 8
& 33.33 & 28.33 & 20.83 & 27.50 \\

& \cellcolor{black!12}\method
& \cellcolor{black!12}Future rollout
& \cellcolor{black!12}8
& \cellcolor{black!12}\textbf{50.56}
& \cellcolor{black!12}\textbf{44.17}
& \cellcolor{black!12}\textbf{29.17}
& \cellcolor{black!12}\textbf{41.30} \\

\bottomrule
\end{tabular}
\end{table}

\textbf{\method outperforms OPSD across both protocols in the answer-available setting.}
Table~\ref{tab:answer-supervised} compares \method with OPSD under two answer-available protocols. 
Following the original OPSD formulation \citep{zhao2026selfdistilled}, the $n=1$ setting uses a reference solution as privileged information. 
In this setting, \method improves the average score from 24.62 to 27.41 on Qwen3-4B and from 37.22 to 44.17 on Qwen3-8B.
For the rollout-based variant, OPSD uses a correct student-generated rollout as privileged information, whereas \method uses a correct future-policy rollout. 
The average score rises from 27.50 to 41.30 on Qwen3-4B and from 48.80 to 64.44 on Qwen3-8B. 
\method improves performance on all three benchmarks in both settings. 
Together, these results show that \method improves OPSD with either source of privileged information, with the largest gains when the future policy provides both the privileged rollout and token-level supervision.

\textbf{\method outperforms OPSD in the answer-free setting using only model-generated training signals.}
Table~\ref{tab:answer-free} compares \method with OPSD in the answer-free
setting, where the training data provide only problem statements and neither gold answers nor reference solutions are available. 
Under this setting, \method improves the average score from 26.76 to 30.00 for Qwen3-4B and from 48.67 to 57.41 for Qwen3-8B, with gains on all three benchmarks. 
These results show that future-policy supervision remains effective when the training signal is constructed entirely from model-generated rollouts. 
Appendix~\ref{app:answer-free-routing} further shows that the future majority is more accurate on routed problems.

\begin{table}[tbp]
\centering
\small
\setlength{\tabcolsep}{8pt}
\renewcommand{\arraystretch}{1.10}
\caption{Main results comparison across mathematical reasoning benchmarks under the answer-free setting. OPSD$^\dagger$ denotes our self-majority variant.}
\label{tab:answer-free}
\begin{tabular}{llrrrr}
\toprule
Model & Method & AIME24 & AIME25 & HMMT25 & Avg. \\
\midrule

\multirow{3}{*}{\textit{Qwen3-8B}}
& Base
& 30.55 & 20.83 & 11.11 & 20.83 \\

\cmidrule(lr){2-6}

& OPSD$^\dagger$
& 61.11 & 52.22 & 32.67 & 48.67 \\

& \cellcolor{black!12}\method
& \cellcolor{black!12}\textbf{72.22}
& \cellcolor{black!12}\textbf{59.72}
& \cellcolor{black!12}\textbf{40.28}
& \cellcolor{black!12}\textbf{57.41} \\

\midrule

\multirow{3}{*}{\textit{Qwen3-4B}}
& Base
& 20.00 & 19.17 & 10.83 & 16.67 \\

\cmidrule(lr){2-6}

& OPSD$^\dagger$
& 35.00 & 27.50 & 17.78 & 26.76 \\

& \cellcolor{black!12}\method
& \cellcolor{black!12}\textbf{40.56}
& \cellcolor{black!12}\textbf{29.72}
& \cellcolor{black!12}\textbf{19.72}
& \cellcolor{black!12}\textbf{30.00} \\

\bottomrule
\end{tabular}
\end{table}

\subsection{Ablation Studies for \method}
We examine the contributions of the future teacher and privileged information
(PI), as well as whether their combination in \method provides benefits beyond
direct OPD. Table~\ref{tab:teacher-pi-ablation} compares base and future
teachers with PI generated by either the student or the future policy, and also
includes a variant that omits PI when the future teacher is selected.
Table~\ref{tab:opd-comparison} further compares \method with direct OPD from
Qwen3-14B and Future Qwen3-4B.

\textbf{Future teacher improves performance under either PI source.}
Replacing the base teacher with the future teacher raises the average score from 27.50 to 34.07 with student PI and from 23.34 to 41.30 with future PI.
Thus, the future teacher contributes beyond generating a rollout for PI: using it as the teacher improves performance even when the PI source is held fixed.
This finding supports the design of \method: temporarily training ahead provides a more effective teacher for the student.

\textbf{Future rollouts provide the most effective PI for the future teacher.}
Without PI, future supervision yields an average score of 23.24. Conditioning the future teacher on a student rollout raises this to 34.07, while using its own rollout further improves it to 41.30. This advantage is specific to the future teacher: the base teacher performs better with a student rollout than with a future rollout (27.50 vs.\ 23.34). 
This reversal suggests a teacher–PI compatibility effect: the base teacher benefits more from student rollouts, whereas the future teacher benefits most from its own rollouts. PI is therefore most effective when its source aligns with the teacher that uses it.
\begin{table}[t]
    \centering
    \small
    \setlength{\tabcolsep}{5pt}
    \renewcommand{\arraystretch}{1.10}
    \caption{Ablation of the teacher and PI in \method under the answer-available setting.
    }
    \label{tab:teacher-pi-ablation}
    \begin{tabular}{llrrrr}
        \toprule
        Teacher & PI & AIME24 & AIME25 & HMMT25 & Avg. \\
        \midrule
        Base   & Student rollout
               & 33.33 & 28.33 & 20.83 & 27.50 \\
        Base   & Future rollout
               & 26.67 & 26.67 & 16.67 & 23.34 \\
        \midrule
        Future & None
               & 28.61 & 25.56 & 15.56 & 23.24 \\
        Future & Student rollout
               & 41.67 & 36.94 & 23.61 & 34.07 \\
        Future & Future rollout
               & \textbf{50.56} & \textbf{44.17}
               & \textbf{29.17} & \textbf{41.30} \\
        \bottomrule
    \end{tabular}
\end{table}

\textbf{Comparison with direct OPD baselines.}
We evaluate direct OPD using Qwen3-14B as an external teacher and the
OPSD-trained Future Qwen3-4B as a self-bootstrapped teacher. The two teachers
have comparable standalone performance, yet their distilled students obtain
average scores of only 19.17 and 24.72, respectively, compared with 41.30 for
\method (Table~\ref{tab:opd-comparison}). Teacher performance and protocol
details are provided in Appendix~\ref{app:external-teacher}.

\begin{table}[t]
    \centering
    \small
    \setlength{\tabcolsep}{3.8pt}
    \renewcommand{\arraystretch}{1.10}
    \caption{Comparison with direct OPD under the answer-available setting.
    All methods train a Qwen3-4B student. Direct OPD uses the listed teacher on
    every training example without PI, whereas \method uses selective future
    supervision with verified future-rollout PI.}
    \label{tab:opd-comparison}
    \begin{tabular}{lllrrrr}
        \toprule
        Method & Teacher & PI & AIME24 & AIME25 & HMMT25 & Avg. \\
        \midrule
        OPD & Qwen3-14B & None
        & 23.61 & 21.67 & 12.22 & 19.17 \\
        OPD & Future Qwen3-4B & None
        & 31.67 & 25.56 & 16.94 & 24.72 \\
        \midrule
        \rowcolor{gray!12}
        \method & Future Qwen3-4B & Future rollout
        & \textbf{50.56} & \textbf{44.17} & \textbf{29.17} & \textbf{41.30} \\
        \bottomrule
    \end{tabular}
\end{table}

\subsection{In-Depth Analysis of Future-Teacher Supervision}
In this section, we examine how the construction and timing of the future teacher affect student performance.

\subsubsection{Effects of Teacher Construction}
\textbf{Teachers constructed by training ahead yield the strongest student performance.}
We compare five teacher configurations: a fixed base policy (\textit{Base}), the current student (\textit{Self}), an exponential moving average of the student (\textit{EMA}), and future teachers constructed from the Base and EMA variants, denoted \textit{Future (Base)} and \textit{Future (EMA)}. 
All variants normally use a correct student rollout as PI; when future supervision is activated, a correct future rollout is used instead.
Further details are provided in Appendix~\ref{app:teacher-source-details}.
Table~\ref{tab:teacher_source} shows that Base, Self, and EMA achieve similar average scores of 27.50, 25.09, and 25.83, respectively. 
Future (Base) raises the score to 41.30, while Future (EMA) achieves the highest score of 52.50. 
These gains across both Base and EMA configurations support our use of training ahead to construct a future teacher.

\subsubsection{Timing of Future-Teacher Supervision}
\textbf{Future supervision is most effective at the start of student training.}
We fix the future teacher to $\pi_F=\pi_{50}$ and apply it during a 25-step window at different stages. All variants start from $\pi_0$ and use the base teacher outside that window.
As shown in Table~\ref{tab:future-window}, applying future supervision at Steps~0--25 yields an average score of 41.30. The score falls to 30.19 when the window moves to Steps~25--50 and to 27.96 at Steps~50--75, approaching the 27.50 obtained without future supervision. Introducing the same future teacher early therefore has a greater effect on the trained student than introducing it later.

\begin{table}[t]
    \centering
    \small
    \caption{Student performance after training with different teacher configurations. Numbers in parentheses denote gains over the initial model's average score of 16.67.}
    \label{tab:teacher_source}

    \begin{tabular}{lcccr@{\hspace{0.35em}}l}
        \toprule
        Teacher source
        & AIME24
        & AIME25
        & HMMT25
        & \multicolumn{2}{c}{Avg. (Gain)} \\
        \midrule

        Base
        & 33.33
        & 28.33
        & 20.83
        & 27.50
        & (+10.83) \\

        Self
        & 29.17
        & 29.17
        & 16.94
        & 25.09
        & (+8.42) \\

        EMA
        & 32.50
        & 28.61
        & 16.39
        & 25.83
        & (+9.16) \\

        Future (Base)
        & \underline{50.56}
        & \underline{44.17}
        & \underline{29.17}
        & \underline{41.30}
        & (+24.63) \\

        Future (EMA)
        & \textbf{66.67}
        & \textbf{55.83}
        & \textbf{35.00}
        & \textbf{52.50}
        & (+35.83) \\

        \bottomrule
    \end{tabular}
\end{table}

\begin{table}[t]
    \centering
    \small
    \setlength{\tabcolsep}{6.5pt}
    \renewcommand{\arraystretch}{1.10}
    \caption{
    Effect of future-supervision timing on policy performance.
    All variants start from $\pi_0$. The fixed future teacher $\pi_{50}$ is used during the indicated 25-step window, and the base teacher is used otherwise.
    }
    \label{tab:future-window}
    \begin{tabular}{lrrrr}
        \toprule
        Future-supervision window
        & AIME24 & AIME25 & HMMT25 & Avg. \\
        \midrule
        None
        & 33.33 & 28.33& 20.83 & 27.50 \\
        \rowcolor{gray!4}
        Steps 50--75
        & 34.44 & 30.28 & 19.17 & 27.96 \\
        \rowcolor{gray!8}
        Steps 25--50
        & 37.78 & 30.56 & 22.22 & 30.19 \\
        \rowcolor{gray!14}
        Steps 0--25
        & \textbf{50.56}& \textbf{44.17} & \textbf{29.17} & \textbf{41.30} \\
        \bottomrule
    \end{tabular}
\end{table}

\subsection{Rationale for Key Design Choices}
In this section, we examine two choices in \method: restarting the student from the base policy and using model-generated trajectories as privileged information.

\subsubsection{Restarting the Student from the Base Policy}
\label{sec:restart-landscape}
Preceding analysis shows that future supervision is most effective early in student training. 
We now examine why \method restarts the student from $\pi_0$ instead of continuing optimization from $\pi_F$.

\begin{figure*}[t]
    \centering
    \includegraphics[width=\textwidth]{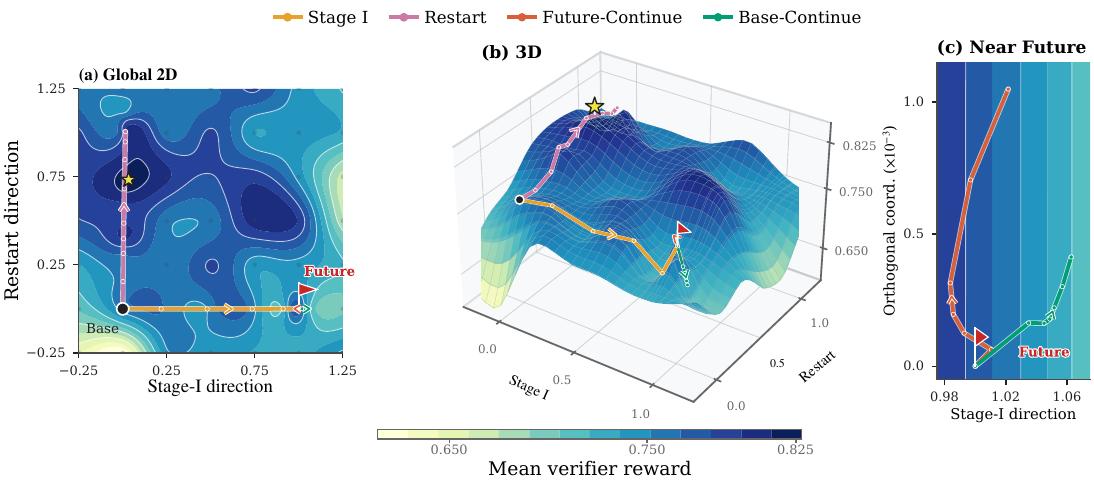}
    \caption{\textbf{Optimization trajectories under restart and continuation.}
    (A) Two-dimensional verifier-reward landscape and optimization trajectories in a shared parameter plane. (B) Corresponding three-dimensional view. (C) Magnified view around $\pi_F$, showing the two continuation trajectories. Darker colors indicate higher reward on the held-out problems.  The red flag marks $\pi_F$, small dots denote evaluated grid models, circular path markers denote evaluated training
    checkpoints, and the star marks the maximum of the interpolated surface.}
    \label{fig:restart-landscape}
\end{figure*}

\textbf{Restarting the student from $\pi_0$ leads to a higher-reward region.}
Figure~\ref{fig:restart-landscape} compares the restarted student with two alternatives that continue from $\pi_F$: one is supervised by the base teacher, and the other by the future teacher.
Following \citet{li2018visualizing}, we project their LoRA checkpoints onto a shared two-dimensional parameter plane and measure mean verifier reward on held-out problems. 
In this projection, both continuation trajectories remain near $\pi_F$, whereas the restarted student moves toward a region with higher reward. Details of the visualization are provided in Appendix~\ref{app:trajectory-landscape}.
These results suggest that using $\pi_F$ to supervise a student restarted from $\pi_0$ can lead to a more effective optimization than continuing training from $\pi_F$.

\subsubsection{Selection of Privileged Information}
\textbf{Model-generated rollouts provide more effective PI than final answers or reference solutions.}
To examine this choice, we fix the base teacher and $n=8$ and compare four forms of PI: a final answer, a dataset-provided reference solution, a correct student rollout, and the same rollout with its boxed answer removed. 
The corresponding prompt templates are provided in Appendix~\ref{app:pi-details}.
Table~\ref{tab:pi-kl-localization} shows that the final answer alone produces the largest KL on answer tokens, but the smallest KL on reasoning tokens and the lowest downstream score of 18.98. 
A correct rollout achieves 27.50. 
Removing its boxed answer retains 93.7\% of its reasoning-token KL and yields a score of 25.83, suggesting that the reasoning trajectory supplies most of the useful information in PI.
The source of that trajectory also matters. Replacing the reference solution with a correct student rollout increases reasoning-token KL from 0.0585 to 0.0746 and downstream performance from 20.83 to 27.50. 
These results motivate the use of rollout-based PI and the stronger $n=8$ student-rollout OPSD baseline in our main comparisons.

\begin{table}[t]
\centering
\small
\setlength{\tabcolsep}{7pt}
\renewcommand{\arraystretch}{1.12}
\caption{
Effect of privileged information on OPSD. All variants use $n=8$ and a fixed base teacher. Performance is averaged across AIME24, AIME25, and HMMT25.
}
\label{tab:pi-kl-localization}
\begin{tabular}{@{}lccc@{}}
\toprule
Teacher PI
& \multicolumn{2}{c}{Raw KL per token}
& Downstream \\
\cmidrule(lr){2-3}
\cmidrule(l){4-4}
& Reasoning
& Answer
& Mean \textsc{Avg@12}$\uparrow$ \\
\midrule
Final answer only
& 0.0221
& \textbf{0.5268}
& 18.98 \\
Reference solution
& 0.0585
& 0.3531
& 20.83 \\
Correct rollout without boxed answer
& 0.0699
& 0.1251
& 25.83 \\
\rowcolor{gray!15}
Correct rollout
& \textbf{0.0746}
& 0.2527
& \textbf{27.50} \\
\bottomrule
\end{tabular}
\end{table}

\subsection{Training Efficiency}
\begin{figure}[t]
    \centering
    \includegraphics[width=0.85\linewidth]{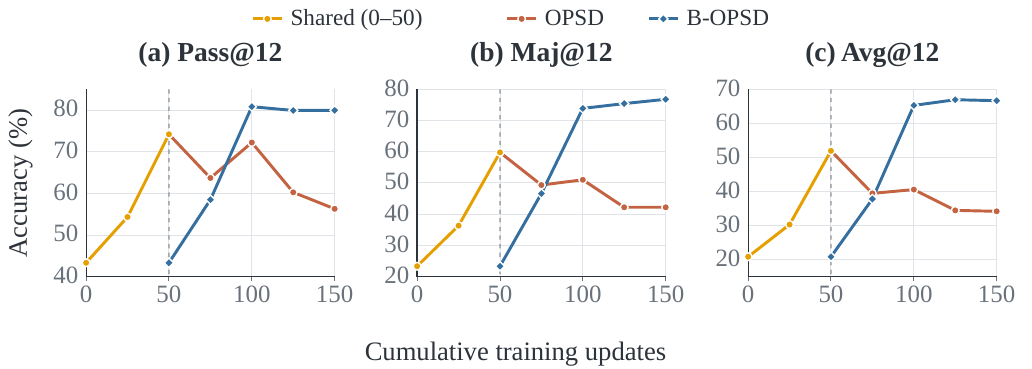}
    \caption{Comparison of AIME25 performance at matched cumulative update budgets for Qwen3-8B.
    The gray line in each panel marks the end of future-policy construction. 
    The \method curve begins there with the student restored to $\pi_0$, while OPSD continues from $\pi_{50}$.}
    \label{fig:compute-dynamics}
\end{figure}

Constructing the future teacher requires 50 optimization updates. 
To determine whether \method's gains are simply due to more training, we compare it with uninterrupted OPSD at the same cumulative update count.
Specifically, after $s$ updates of the restarted student, \method is compared with OPSD at update $50+s$, counting the 50 updates used to construct the future teacher.

\textbf{\method reaches higher peak performance than vanilla OPSD within the same cumulative update budget.}
Figure~\ref{fig:compute-dynamics} compares their evaluation performance on AIME25 using Qwen3-8B.
After the first 50 updates, OPSD continues from $\pi_{50}$, whereas \method restarts from $\pi_0$ and learns from $\pi_{50}$ as its future teacher.
The restarted student initially performs worse, but subsequently surpasses uninterrupted OPSD and achieves higher peak scores on all three evaluation metrics.
This result indicates that the gain cannot be explained solely by the additional updates used to construct the future teacher.
Results for the remaining benchmarks and metrics are reported in Appendix~\ref{app:additional-results}.

\section{Conclusion}
Our study shows that effective OPSD depends jointly on the capability of the self-teacher and the privileged information it interprets.
We proposed Bootstrapped On-Policy Self-Distillation (\method), which demonstrates that optimization progress can be turned into stronger supervision for an earlier policy state, with future-generated trajectories providing the greatest benefit when paired with the future teacher.
Our analyses further reveal that this supervision is most effective early after restart and can lead to better learning trajectories than simply continuing from the future checkpoint.
Overall, our findings position optimization progress as a reusable source of supervision, broadening self-improvement beyond simply accumulating better parameters.

\bibliography{iclr2027_conference}
\bibliographystyle{iclr2027_conference}

\appendix

\section{Algorithmic Details}
This section presents the complete procedures of \method under the answer-available and answer-free settings. 
Both begin with temporary OPSD updates to construct a future teacher, then restore the student to its initial state and train it using supervision from that teacher. 
The two algorithms differ in how they select privileged information and determine when to use the future teacher: the answer-available setting uses known correct answers, whereas the answer-free setting relies on agreement among generated answers.

\begin{algorithm}[t]
\caption{\method in the Answer-Available Setting}
\label{alg:lookahead_opsd_answer}
\begin{algorithmic}[1]
\Require Base policy $\pi_0$; dataset
$\mathcal{D}=\{(x,a,y^\star)\}$, where $a$ is the gold answer
and $y^\star$ is the gold reasoning trace; lookahead steps $K$;
sample size $N$

\Statex \hspace{-\algorithmicindent}\textit{Notation:}
$\operatorname{Distill}(\pi_S,\pi_T;y_S,y_P)$ updates only
$\pi_S$ using Eq.~\ref{eq:bootstrapped_opsd}, with frozen teacher
$\pi_T$, student trajectory $y_S$, and privileged trajectory $y_P$.

\State \textbf{Stage I: Construct the future self}
\State $\pi_F\gets\pi_0$
\For{$k=1,\ldots,K$}
    \State Sample $(x,a,y^\star)\sim\mathcal{D}$
    \State Sample $\mathcal{Y}
        =\{y^{(i)}\sim\pi_F(\cdot\mid x)\}_{i=1}^{N}$
    \If{$\mathcal{Y}$ contains an incorrect trajectory}
        \State Select an incorrect $y^{-}\in\mathcal{Y}$
            \Comment{On-policy student trajectory}
        \State Select a correct $y^{+}\in\mathcal{Y}$ if available;
            otherwise set $y^{+}\gets y^\star$
        \State $\pi_F\gets
            \operatorname{Distill}(\pi_F,\pi_0;y^{-},y^{+})$
            \Comment{Base policy as teacher}
    \EndIf
\EndFor
\State Freeze $\pi_F$

\Statex
\State \textbf{Stage II: Restart student and learn from the future self}
\State $\pi_S\gets\pi_0$
\For{each training update}
    \State Sample $(x,a,y^\star)\sim\mathcal{D}$
    \State Sample $\mathcal{Y}_S
        =\{y_S^{(i)}\sim\pi_S(\cdot\mid x)\}_{i=1}^{N}$
    \If{$\mathcal{Y}_S$ contains an incorrect trajectory}
        \State Select an incorrect $y_S^{-}\in\mathcal{Y}_S$
            \Comment{On-policy student trajectory}
        \State Sample $y_F\sim\pi_F(\cdot\mid x)$
        \If{$y_F$ is correct}
            \State $\pi_S\gets
                \operatorname{Distill}(\pi_S,\pi_F;y_S^{-},y_F)$
                \Comment{Future-self pathway}
        \Else
            \State Select a correct $y_S^{+}\in\mathcal{Y}_S$
                if available; otherwise set $y_S^{+}\gets y^\star$
            \State $\pi_S\gets
                \operatorname{Distill}(\pi_S,\pi_0;y_S^{-},y_S^{+})$
                \Comment{Standard OPSD pathway}
        \EndIf
    \EndIf
\EndFor
\State \Return $\pi_S$
\end{algorithmic}
\end{algorithm}

\begin{algorithm}[t]
\caption{\method in the Answer-Free Setting}
\label{alg:lookahead_opsd_answer_free}
\begin{algorithmic}[1]
\Require Base policy $\pi_0$; unlabeled dataset
$\mathcal{D}=\{x\}$; lookahead steps $K$; sample size $N$

\Statex \hspace{-\algorithmicindent}\textit{Notation:}
$\operatorname{Distill}(\pi_S,\pi_T;y_S,y_P)$ updates only
$\pi_S$ using Eq.~\eqref{eq:bootstrapped_opsd}, with frozen teacher
$\pi_T$, student trajectory $y_S$, and privileged trajectory $y_P$.
\Statex \hspace{-\algorithmicindent}
For a rollout set, the pseudo-label is its most frequent final answer;
its support is the number of rollouts producing that answer.

\State \textbf{Stage I: Construct the future self}
\State $\pi_F\gets\pi_0$
\For{$k=1,\ldots,K$}
    \State Sample a problem $x\sim\mathcal{D}$
    \State Sample $\mathcal{Y}
        =\{y^{(i)}\sim\pi_F(\cdot\mid x)\}_{i=1}^{N}$
    \State Obtain pseudo-label $\hat{a}$ from $\mathcal{Y}$
    \If{$\mathcal{Y}$ contains a trajectory inconsistent with $\hat{a}$}
        \State Select $y^{+}\in\mathcal{Y}$ consistent with $\hat{a}$
            \Comment{Privileged trajectory}
        \State Select $y^{-}\in\mathcal{Y}$ inconsistent with $\hat{a}$
            \Comment{Student trajectory}
        \State $\pi_F\gets
            \operatorname{Distill}(\pi_F,\pi_0;y^{-},y^{+})$
            \Comment{Base policy as teacher}
    \EndIf
\EndFor
\State Freeze $\pi_F$

\Statex
\State \textbf{Stage II: Restart student and learn from the future self}
\State $\pi_S\gets\pi_0$
\For{each training update}
    \State Sample a problem $x\sim\mathcal{D}$
    \State Sample $\mathcal{Y}_S
        =\{y_S^{(i)}\sim\pi_S(\cdot\mid x)\}_{i=1}^{N}$
    \State Obtain pseudo-label $\hat{a}_S$ and support $c_S$
        from $\mathcal{Y}_S$

    \If{$\mathcal{Y}_S$ contains no trajectory inconsistent with $\hat{a}_S$}
        \State \textbf{continue}
        \Comment{No student trajectory for distillation}
    \ElsIf{$c_S > N/2$}
        \State Select $y_S^{+}\in\mathcal{Y}_S$ consistent with $\hat{a}_S$
        \State Select $y_S^{-}\in\mathcal{Y}_S$ inconsistent with $\hat{a}_S$
        \State $\pi_S\gets
            \operatorname{Distill}(\pi_S,\pi_0;y_S^{-},y_S^{+})$
            \Comment{Standard OPSD pathway}
    \Else
        \State Sample $\mathcal{Y}_F
            =\{y_F^{(i)}\sim\pi_F(\cdot\mid x)\}_{i=1}^{N}$
        \State Obtain pseudo-label $\hat{a}_F$ and support $c_F$
            from $\mathcal{Y}_F$
        \If{$c_F > N/2$}
            \State Select $y_F^{+}\in\mathcal{Y}_F$
                consistent with $\hat{a}_F$
                \Comment{Privileged trajectory}
            \State Select $y_S^{-}\in\mathcal{Y}_S$
                inconsistent with $\hat{a}_F$
                \Comment{Student trajectory}
            \State $\pi_S\gets
                \operatorname{Distill}(\pi_S,\pi_F;y_S^{-},y_F^{+})$
                \Comment{Future-self pathway}
        \Else
            \State Select $y_S^{+}\in\mathcal{Y}_S$
                consistent with $\hat{a}_S$
            \State Select $y_S^{-}\in\mathcal{Y}_S$
                inconsistent with $\hat{a}_S$
            \State $\pi_S\gets
                \operatorname{Distill}(\pi_S,\pi_0;y_S^{-},y_S^{+})$
                \Comment{Standard OPSD fallback}
        \EndIf
    \EndIf
\EndFor
\State \Return $\pi_S$
\end{algorithmic}
\end{algorithm}

\section{Implementation Details}
\label{app:implementation}
The reproducible codes will be published upon the acceptance of the paper.

We use Qwen3-4B and Qwen3-8B in non-thinking mode. The original OPSD setting
uses $n=1$: each update samples one student trajectory for each of 32
problems, and the dataset reference solution is provided as privileged
information to the teacher.

Our $n=8$ OPSD baseline samples eight stochastic student trajectories for each
problem, giving 256 candidate trajectories per update. These candidates are
reduced to one training row per problem. In the answer-available setting, we
use the longest parseable gold-correct rollout as PI and the longest parseable
gold-wrong rollout as the student trajectory. If all student rollouts are
incorrect, the dataset reference solution is used as PI.

The answer-available Future experiments use the same $n=8$ student rollouts.
During the Future-supervision window, the frozen future policy produces one
additional greedy rollout for each active problem, with temperature 0,
top-$p$ 1.0, and top-$k$ disabled. A verified correct rollout activates the
future teacher and is used as PI; otherwise, the example follows the Base
OPSD pathway. Training rollouts use temperature 1.1, top-$p$ 0.95, top-$k$ 20,
and a maximum length of 4,096 tokens. Evaluation uses 12 sampled responses per
problem with temperature 1.0, top-$p$ 0.8, no top-$k$ truncation, and a maximum
response length of 38,912 tokens. The evaluation context length is 40,960
tokens.

\paragraph{Parameter-efficient training.}
We train LoRA adapters with rank 64 and scaling factor 128. The adapters are
inserted into the query, key, value, and output projections of self-attention,
as well as the gate, up, and down projections of the feed-forward blocks. We
use AdamW with learning rate $5\times 10^{-6}$, betas $(0.9,0.999)$, zero
weight decay, a constant learning-rate schedule, and gradient-norm clipping at
0.1. We use one optimizer epoch per generated batch and a per-GPU actor micro
batch size of one. No entropy bonus, PPO importance
sampling ratio, or separate old-policy correction is used in the OPSD
objective.

\paragraph{Token-level distillation.}
For every active response position, the teacher and student distributions are
computed over the full vocabulary. Only response tokens contribute to the
loss; prompt and padding positions are masked out. 

\begin{figure*}[t]
    \centering
    \includegraphics[width=.85\textwidth]{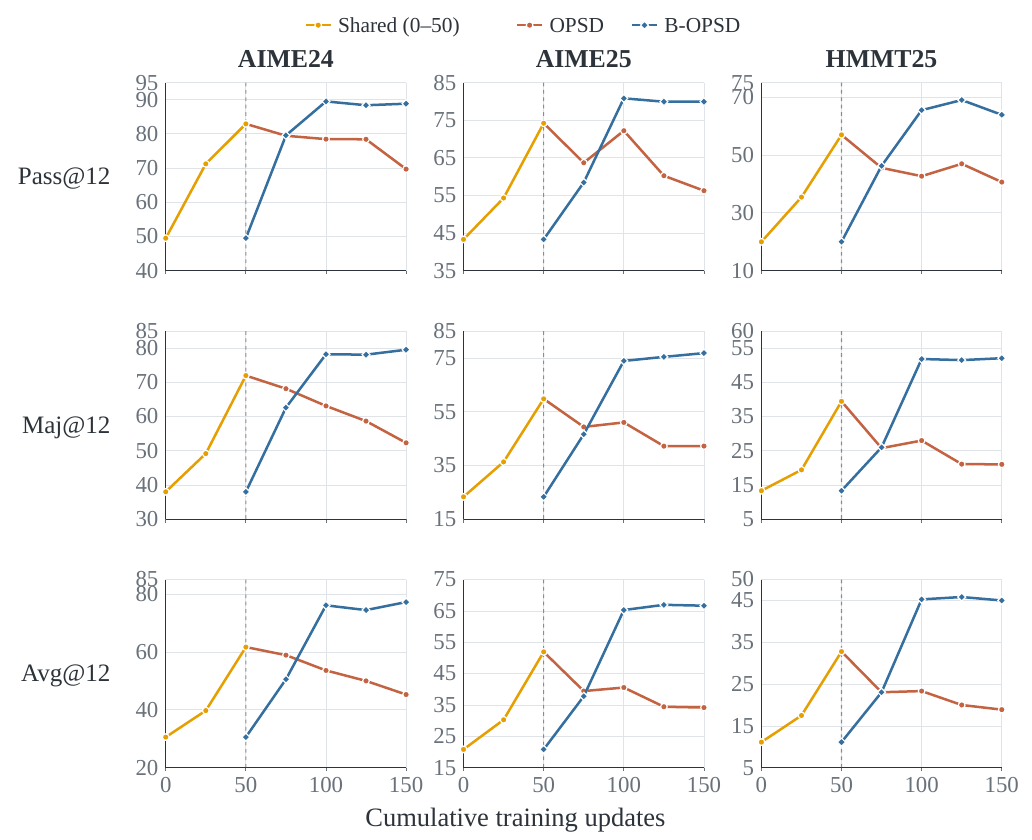}
    \caption{\textbf{Matched-budget training dynamics across benchmarks and
    metrics.} Columns show AIME24, AIME25, and HMMT25; rows show
    \textsc{Pass@12}, \textsc{Maj@12}, and \textsc{Avg@12}. The gray dashed
    line marks the end of the 50 lookahead updates. The curves are compared
    at shared cumulative budgets of 50, 75, 100, 125, and 150 updates.}
    \label{fig:compute-dynamics-full}
\end{figure*}
\paragraph{Trajectory selection and verification.}
For the answer-available $n=8$ baseline, the longest parseable gold-correct
student rollout is preferred as PI and the longest parseable gold-wrong rollout
is used as the student trajectory. On all-wrong groups, the PI candidate is
excluded from student-target selection and the dataset reference solution is
used as the PI fallback. For the $n=1$ reproduction, the single
student rollout is used as the on-policy trajectory and the dataset reference
solution remains the PI; no multi-rollout selection is performed.

For answer-free training, the final answers extracted from the sampled
\verb|\boxed{}| expressions are aggregated by plurality vote. The longest
rollout supporting the selected answer is used as privileged information, and
the longest rollout disagreeing with the selected answer is used as the
student trajectory. The future pathway is queried only when the student group
does not have a strict majority; it is accepted only when the future group has
a strict majority and an eligible student trajectory disagrees with the future
majority answer. If either condition fails, the update falls back to standard
OPSD. A response is marked invalid when generation reaches the maximum length
without an end-of-sequence token. The final answer is the last parseable
\verb|\boxed{}| expression, and mathematical equivalence is checked with a
symbolic verifier followed by normalized string matching when parsing fails.

\paragraph{Data and evaluation protocol.}
Training uses the OPSD split of OpenThoughts-Math-30K, containing 29,434
problems. The evaluation suite consists of AIME 2024, AIME 2025, and HMMT
February 2025. We report average accuracy over sampled responses, pass rate,
and majority-vote accuracy. The complete prompt templates and the additional
teacher-source protocols are given in Appendices~\ref{app:pi-details}
and~\ref{app:teacher-source-details}.

\section{Additional Results}
\label{app:additional-results}
Figure~\ref{fig:compute-dynamics-full} reports the complete matched-budget
comparison across all benchmarks and metrics.

\section{Privileged-Information Prompt Templates}
\label{app:pi-details}

PI is provided only to the teacher. The student receives the original problem
followed by the instruction to reason step by step and place the final answer
in \verb|\boxed{}|. We use three prompt templates according to the content of
the PI. The answer-only template supplies only the verified final answer and is
used in the PI-content ablation in Table~\ref{tab:pi-kl-localization}. The
reference-solution template is used for dataset-provided solutions. The
candidate-solution template is used for verified student and future-policy
rollouts. Braced fields below are replaced with the corresponding problem,
answer, or reasoning trajectory.

\begin{promptbox}{Answer-Only PI Prompt}
Problem: \{problem\}

\smallskip
Here is the verified final answer to this problem:\newline
=== Final Answer Begin ===\newline
\{answer\}\newline
=== Final Answer End ===

\smallskip
After reading the verified final answer above, make sure you understand what
answer your reasoning should derive---do not merely copy or restate it. Now,
using your own words and independent reasoning, derive the same final answer
to the problem above. Think step by step, explore different approaches, and
don't be afraid to backtrack or reconsider if something doesn't work out:

\smallskip
Please reason step by step, and put your final answer within
\textbackslash boxed\{\}.
\end{promptbox}

\begin{promptbox}{Reference-Solution PI Prompt}
Problem: \{problem\}

\smallskip
Here is a reference solution to this problem:\newline
=== Reference Solution Begin ===\newline
\{solution\}\newline
=== Reference Solution End ===

\smallskip
After reading the reference solution above, make sure you truly understand the
reasoning behind each step---do not copy or paraphrase it. Now, using your own
words and independent reasoning, derive the same final answer to the problem
above. Think step by step, explore different approaches, and don't be afraid to
backtrack or reconsider if something doesn't work out:

\smallskip
Please reason step by step, and put your final answer within
\textbackslash boxed\{\}.
\end{promptbox}

\begin{promptbox}{Model-Generated Rollout PI Prompt}
Problem: \{problem\}

\smallskip
Here is a candidate solution to this problem:\newline
=== Candidate Solution Begin ===\newline
\{solution\}\newline
=== Candidate Solution End ===

\smallskip
After reading the candidate solution above, make sure you truly understand the
reasoning behind each step---do not copy or paraphrase it. Now, using your own
words and independent reasoning, derive the final answer to the problem above.
Think step by step, explore different approaches, and don't be afraid to
backtrack or reconsider if something doesn't work out:

\smallskip
Please reason step by step, and put your final answer within
\textbackslash boxed\{\}.
\end{promptbox}

\section{Teacher-Source Protocol}
\label{app:teacher-source-details}

This section specifies the Qwen3-4B teacher constructions used in
Table~\ref{tab:teacher_source}. All variants use the same training data,
rollout-based PI selection, optimizer, and evaluation protocol; they differ in
the policy that produces the token-level targets and, for the future variants,
the short period during which future supervision is enabled.

\begin{table*}[t]
    \centering
    \small
    \setlength{\tabcolsep}{5pt}
    \renewcommand{\arraystretch}{1.12}
    \caption{Teacher constructions for the Qwen3-4B teacher-source comparison.
    Both future variants restart the student from the base checkpoint and keep
    the future policy frozen. Future supervision is applied conditionally during
    the first 25 restarted-student updates; subsequent updates use the
    corresponding Base or EMA teacher.}
    \label{tab:teacher-source-protocol}
    \begin{tabular}{lp{0.25\textwidth}p{0.25\textwidth}p{0.29\textwidth}}
        \toprule
        Variant & Token-level teacher & Future construction & Restarted-student schedule \\
        \midrule
        Base
        & Frozen initial policy $\pi_0$
        & --
        & The fixed teacher is used throughout training. \\

        Self
        & Current student policy $\pi_S$
        & --
        & The teacher is synchronized with the current student at every update. \\

        EMA
        & Exponential moving average $\bar{\pi}_S$
        & --
        & The EMA teacher is updated after each student update with rate $0.05$. \\

        Future (Base)
        & Frozen future policy when selected; otherwise Base
        & The checkpoint after 50 OPSD updates with the Base teacher
        & Future supervision is enabled for updates 1--25; Base is used as the
        fallback and for all later updates. \\

        Future (EMA)
        & Frozen future policy when selected; otherwise EMA
        & The checkpoint after 50 OPSD updates with the EMA teacher
        & Future supervision is enabled for updates 1--25; EMA is used as the
        fallback and for all later updates. \\
        \bottomrule
    \end{tabular}
\end{table*}

For Base, Self, and EMA, a verified correct student rollout is used as PI when
available, with the reference solution as a fallback. During the initial
future-supervision phase, the frozen future policy generates one greedy rollout
when the student rollout is incorrect. A verified correct future rollout
activates the future teacher and also serves as its PI; otherwise, training
follows the corresponding Base or EMA pathway. Thus, Future (EMA) differs from
EMA only through the first 25 restarted-student updates, after which both use
the same EMA teacher construction.

\section{Direct OPD with Qwen3-14B and the Future Teacher}
\label{app:external-teacher}

We evaluate direct OPD with two frozen teachers: Qwen3-14B and the Future
Qwen3-4B checkpoint obtained after 50 OPSD updates.

\paragraph{OPD protocol.}
Both variants train a Qwen3-4B student on the same on-policy responses used in
the corresponding OPSD setting. The teacher is conditioned only on the
original problem and provides token-level targets for every training example,
without PI or selective routing. We optimize the full-vocabulary
forward KL $D_{\mathrm{KL}}(p_T\|p_S)$ at each response-token position. All remaining rollout, optimization, and
evaluation settings follow Appendix~\ref{app:implementation}.

\paragraph{Teacher capability.}
Table~\ref{tab:external-teacher-capability} compares their standalone
performance. Future Qwen3-4B and Qwen3-14B obtain similar average scores of
23.89 and 23.05, respectively, although their results differ across individual
benchmarks. Both outperform the initial Qwen3-4B checkpoint on average.

\begin{table}[t]
    \centering
    \small
    \setlength{\tabcolsep}{7pt}
    \renewcommand{\arraystretch}{1.08}
    \caption{Standalone performance of the Qwen3-4B initialization and the two
    teachers used for direct OPD. Future Qwen3-4B is obtained after 50
    lookahead OPSD updates.}
    \label{tab:external-teacher-capability}
    \begin{tabular}{lcccc}
        \toprule
        Model & AIME24 & AIME25 & HMMT25 & Avg. \\
        \midrule
        Qwen3-4B & 20.00 & 19.17 & 10.83 & 16.67 \\
        Future Qwen3-4B & 27.78 & \textbf{26.11} & \textbf{17.78} & \textbf{23.89} \\
        Qwen3-14B & \textbf{30.00} & 25.83 & 13.33 & 23.05 \\
        \bottomrule
    \end{tabular}
\end{table}

\paragraph{Distillation results.}
The corresponding distillation results are reported in
Table~\ref{tab:opd-comparison}. Despite their comparable standalone averages,
direct OPD from Qwen3-14B reaches 19.17 and direct OPD from Future Qwen3-4B
reaches 24.72, whereas \method reaches 41.30. These results highlight two
advantages of \method: it outperforms OPD from the larger Qwen3-14B teacher
without relying on an external model, and it substantially improves over
direct distillation from the same Future Qwen3-4B teacher. 

\section{Future-Self Majority Accuracy on Routed Problems}
\label{app:answer-free-routing}

We further examine the answer-free problems for which the future-self pathway
is selected. Gold answers are used only for this post-hoc analysis. As shown in
Table~\ref{tab:answer-free-routing-accuracy}, the majority-voted answer from the
future policy is more accurate than that from the restarted student for both
model sizes. This confirms that future routing provides a more reliable
pseudo-label on the problems where it is used.

\begin{table}[t]
    \centering
    \small
    \setlength{\tabcolsep}{8pt}
    \caption{Majority-vote accuracy on answer-free training problems routed to
    the future-self pathway. Gold answers are used only for evaluation.}
    \label{tab:answer-free-routing-accuracy}
    \begin{tabular}{lcccc}
        \toprule
        Model & Routed problems & Student Maj. & Future Maj. & Gain \\
        \midrule
        Qwen3-4B & 72 & 65.3 & \textbf{70.8} & +5.6 \\
        Qwen3-8B & 75 & 66.7 & \textbf{73.3} & +6.7 \\
        \bottomrule
    \end{tabular}
\end{table}

\section{Verifier-Reward Landscape Construction}
\label{app:trajectory-landscape}

We implement the verifier-reward landscape as a two-dimensional optimization-landscape
visualization over Qwen3-4B LoRA checkpoints.

\paragraph{Parameter plane.}
For each adapted module $l$, we represent a LoRA checkpoint by its effective
weight update
$\Delta W_l=(\alpha_l/r_l)B_lA_l$.
The base model is the origin.  The first direction $u$ is the effective update
from Base to the Stage-I Future checkpoint $\pi_F=\pi_{50}$.  The second
direction is the Base-to-Restart update at step 100 after removing its
projection onto $u$; we rescale this orthogonal component to match
$\lVert u\rVert$.  A coordinate $(x,y)$ therefore represents the effective
update $x u+y v$.  Saved checkpoints are projected onto the same plane using
Frobenius inner products over the adapted modules.  We estimate these inner
products using 2,048 uniformly sampled effective-update coordinates per module,
averaged over three fixed sketch seeds.  Each grid point is instantiated as a
rank-128 LoRA adapter by combining the two rank-64 direction adapters.

\paragraph{Compared trajectories.}
Stage~I trains from $\pi_0$ for 50 updates to obtain $\pi_F$.
\textit{Restart} returns to $\pi_0$ and enables supervision from $\pi_F$
during updates 1--25.  Both continuation variants resume from $\pi_F$:
\textit{Future-Continue} enables the same future supervision during updates
51--75, whereas Base-Continue keeps the base teacher throughout.
Restart and Future-Continue use the base-teacher pathway outside their
respective future-supervision windows.

\paragraph{Reward surface and trajectories.}
We evaluate a $7\times7$ grid with both coordinates ranging from $-0.25$ to
$1.25$ in increments of $0.25$.  Each point is evaluated on the same 48 math
problems drawn after excluding examples used by the Stage-I, Continue, and
Restart runs, with four non-thinking rollouts per problem.  We use temperature
$1.1$, top-$p=1.0$, and a maximum response length of 8192 tokens.  Mean
verifier reward is the fraction of the resulting 192 responses with a correct
final answer.

The plotted paths comprise five Stage-I checkpoints, six checkpoints from each
continuation branch, and eleven Restart checkpoints.  We evaluate all 49 grid
points and 28 checkpoints under the same protocol, then use a smoothed
thin-plate-spline interpolation to visualize the reward surface between the
evaluated locations.  The markers on the paths denote the directly evaluated
checkpoints; the interpolation is used only for the background surface.

\end{document}